\documentclass[default,iicol]{sn-jnl}

\usepackage{graphicx}%
\usepackage{multirow}%
\usepackage{amsmath,amssymb,amsfonts}%
\usepackage{amsthm}%
\usepackage{mathrsfs}%
\usepackage[title]{appendix}%
\usepackage{xcolor}%
\usepackage{textcomp}%
\usepackage{manyfoot}%
\usepackage{booktabs}%
\usepackage{algorithm}%
\usepackage{algorithmicx}%
\usepackage{algpseudocode}%
\usepackage{listings}%
\usepackage{float}
\usepackage{epsfig}
\usepackage{booktabs}
\usepackage{multirow}
\usepackage{arydshln}
\usepackage{subcaption}
\usepackage{float}

\usepackage{acronym}
\acrodef{ITSR}{Image Transformation Sequence Retrieval}
\acrodef{MCTS}{Monte-Carlo Tree Search}

\begin{document}

\title[Image Transformation Sequence Retrieval with General Reinforcement Learning]{Image Transformation Sequence Retrieval with General Reinforcement Learning}


\author[1]{\fnm{Enrique} \sur{Mas-Candela}}\email{emc89@alu.ua.es}

\author*[1]{\fnm{Antonio} \sur{Rios-Vila}}\email{arios@dlsi.ua.es}

\author[1]{\fnm{Jorge} \sur{Calvo-Zaragoza}}\email{jcalvo@dlsi.ua.es}

\affil*[1]{\orgdiv{U.I for Computing Research}, \orgname{University of Alicante}, \country{Spain}}


\abstract{In this work, the novel Image Transformation Sequence Retrieval (ITSR) task is presented, in which a model must retrieve the sequence of transformations between two given images that act as source and target, respectively. Given certain characteristics of the challenge such as the multiplicity of a correct sequence or the correlation between consecutive steps of the process, we propose a solution to ITSR using a general model-based Reinforcement Learning such as Monte Carlo Tree Search (MCTS), which is combined with a deep neural network. Our experiments provide a benchmark in both synthetic and real domains, where the proposed approach is compared with supervised training. The results report that a model trained with MCTS is able to outperform its supervised counterpart in both the simplest and the most complex cases. Our work draws interesting conclusions about the nature of ITSR and its associated challenges.}




\maketitle

\section{Introduction}
On the way to Artificial General Intelligence (AGI), benchmarks such as the Abstraction and Reasoning Corpus (ARC) have been released \cite{chollet2019measure}. This challenge consists in presenting to the \emph{intelligent} entity (human or machine) a series of demonstration examples, in the form of input-output grids. The entity is then given a small number of test examples and asked to generate the corresponding output grids.

The ARC challenge showcases that there are plenty of tasks that a human could \emph{easily} solve but that are extremely complex for the dominant models of the machine learning landscape. This is commonly attributed to the fact that such problems are based on priors that are natural to the human psyche, yet difficult to transfer specifically to machine learning protocols \cite{hernandez2017measure}. Solutions to the corresponding competition over ARC resulted in rather ad-hoc approaches.
\footnote{https://www.kaggle.com/competitions/abstraction-and-reasoning-challenge}

\begin{figure*}[ht!]
    \centering

    \begin{subfigure}[b]{\textwidth}
        \includegraphics[width=\textwidth]{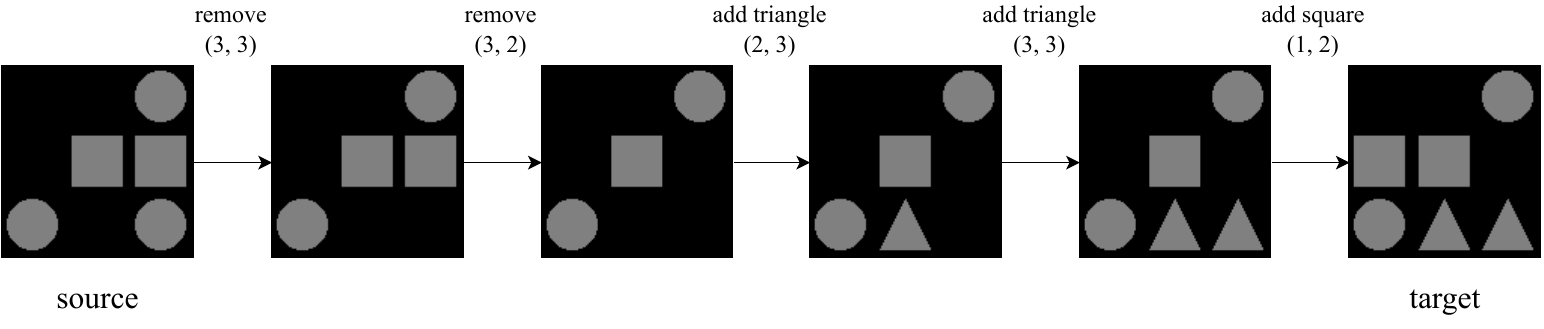}
        \caption{Example of transformation sequence in a toy task where simple objects can be added or removed from a grid.}
        \label{fig:problem-example-toy}
    \end{subfigure}
    \hfill
    \begin{subfigure}[b]{\textwidth}
        \includegraphics[width=\textwidth]{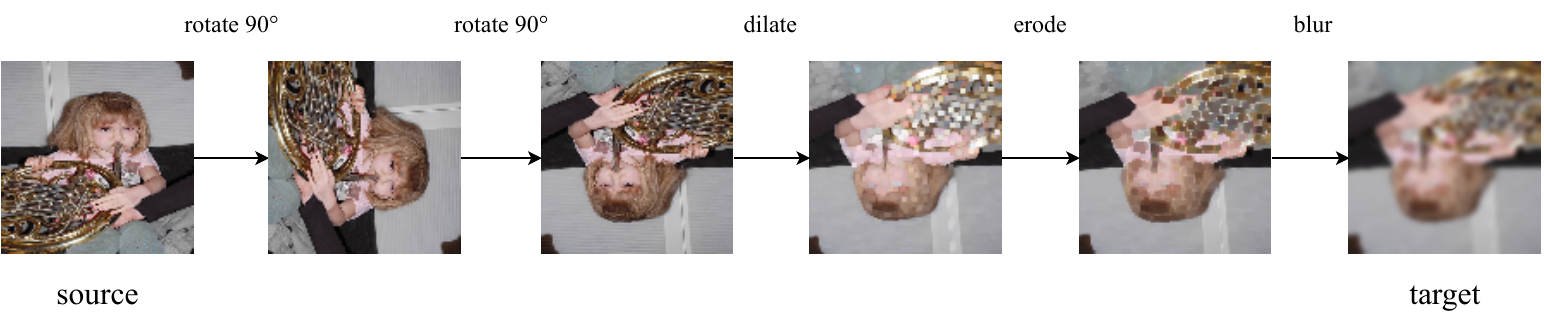}
        \caption{Example of transformation sequence with classic image-processing operators on the Imagenette dataset.}
    \end{subfigure}
    
    \caption{Example of transformation sequences applied in two different scenarios. Given the source and target images, the \ac{ITSR} task aims to retrieve the sequence of transformations applied}
    \label{fig:problem-example}
\end{figure*}

In a reasonable effort towards similar goals, in this work we take a step back to establish a challenge of analogous nature but closer to what neural networks can currently address. Here, we deal with the problem of \ac{ITSR}. Given a source image, \ac{ITSR} seeks to retrieve the sequence of transformations (from a known and finite set) that converts such source into a given target image (see Fig.~\ref{fig:problem-example}). Despite this simple and human-friendly formulation, \ac{ITSR} hides a number of interesting challenges for computer vision and pattern recognition: relational reasoning, abstraction, flexibility, and generality. 

It should be noted that \ac{ITSR} explicitly considers that the solution is represented as a sequence of transformations, unlike other program synthesis challenges in which the solution may be represented as a syntax tree, driven (or not) by a domain-specific language grammar. In addition, depending on the nature of the transformations involved, there may be more than one sequence that effectively converts the source to the target. That is why the problem refers to \emph{retrieval}, as opposed to \emph{prediction}.

Furthermore, in this work, we provide a first attempt to \ac{ITSR} by following a Sequential Decision-Making approach. For each source and target image pair, the model is asked to choose which transformation to apply next, so that the former gets closer to the latter. Since the dynamics of the problem are known, we address the \ac{ITSR} challenge through model-based Reinforcement Learning (RL). Following the example of AlphaZero \cite{silver19alphazero}, we propose a general RL strategy based on \ac{MCTS} combined with deep learning. Our experiments show how this approach provides a greater performance if compared with models trained in a supervised setting. 

The contributions of this work are: i) establish \ac{ITSR} as a novel computer vision and pattern recognition task, easy to formulate but with interesting challenges for the state of the art, ii) a solution to the problem based on the combination of deep learning with model-based RL strategies, and iii) a benchmark both in terms of data and baseline results, that serves as a reference for future approaches.

\section{Related works}
To the best of our knowledge, this is the first time that \ac{ITSR} is considered. However, some similarities with other research fields can be considered.

\subsection{Image forensics.} The research field of image forensics has approached similar challenges of image transformation detection and retrieval. Most of the literature resorts to traditional computer vision algorithms for detecting transformations on images~\cite{feng2012,kang2013}. Neural-specific approaches are also found, where a universal method for detecting different transformations on images is proposed, both with large databases \cite{bayar2018} and in few-shot scenarios \cite{mazumdar2020}. Despite the advances produced by these efforts, all of the image forensics literature is framed in single-transformation classification solutions. The \ac{ITSR} problem consists of a multi-stage edition process where a sequence of transformations has to be retrieved from input-output image pairs, which adds a new level of complexity that is yet unexplored.

\subsection{Neural Program Synthesis.} The automatic generation of programs has been typically addressed by Neural Program Synthesis since the early days of artificial intelligence \cite{Waldinger69}. Similarly to \ac{ITSR}, in this research field, an explicit program must be generated to fulfill a specification given by examples. The emergence of deep learning brought great advances, since they solved some of the limitations of traditional methods \cite{Johnson2017, Parisotto2017, Delvin2017}. Most of these solutions approach this problem via supervised learning, like the recent GitHub Copilot tool \cite{Chen2021b}, which generates programs by taking text description inputs \cite{Brown2020}. The use of RL approaches has also been proposed and demonstrated to work in this area \cite{Bunel2018, Liang2022}. 
Despite these advances, all the proposed solutions are constrained to the natural language processing domain. However, there are program functionality examples that can be shown without text descriptions, such as image edition---where only a source image and its desired target are required. This work aims to fill this domain gap and expand the applicability of this field to the image domain. 

\section{Problem formulation}
\label{sec:definition}
Let $X$ be the image space and $T$ be the set of transformations (functions) that can alter an image in the form of $t : X \rightarrow X\text{,}~t \in T$. These image-transformation functions can be concatenated as a $k$-length sequence $\mathbf{t} = (t_1, t_2, t_3, \ldots , t_k) \in \mathbf{T}^{k}$. Similarly, we can define a function $f(x, \mathbf{t}): \mathcal{X} \rightarrow \mathcal{X}$ that applies this transformation sequence to an input image in order: $f(x, \mathbf{t}) =  t_k(\cdots t_2(t_1(x)))$.

Given a pair of images $(x, x'), x,x' \in X$, \ac{ITSR} seeks for a transformation sequence $\mathbf{t}$ that satisfies $f(x, \mathbf{t}) = x'$. In general, we assume that $\exists\mathbf{t}\in\mathcal{T}^{*}:f(x,\mathbf{t}) = x'$. 
An approach to \ac{ITSR} must estimate a function $M: \mathcal{X} \times \mathcal{X} \rightarrow \mathcal{T}^{*}$ that maps $(x, x')$ into $\mathbf{t}$.

It is worth noting that \ac{ITSR} solutions do not have to be unique but there can be more than one sequence that satisfies the success condition.
This is because there might be image transformation operations that can be swapped and yet produce the same result. 
A simple case is a two-element transformation sequence $\mathbf{t} = \{t_1,t_2\}$, where $t_1$ stands for a vertical flip and $t_2$ for an image inversion. In this particular case, $t_1(t_2(x)) = t_2(t_1(x))$ is satisfied. This can happen at any position of an edition sequence. Consequently, given $(x,x')$, any sequence of the set $\{ \mathbf{t} \in \mathbf{T}^{*} : f(x, \mathbf{t}) = x' \}$ will be considered as a valid solution.\footnote{Note that, contrary to this assumption, one could consider a minimum-length sequence or the one with the lowest cost depending on the operations involved.}

\section{Methodology}
Given that the output of \ac{ITSR} is a sequence, existing image-to-sequence endeavours suggest the use of recurrent neural networks \cite{Horchreiter1997} or Transformers \cite{Vaswani2017} with supervised training. Here, however, multiple sequences of transformations might be a valid solution for the same input pair. We will see later that this condition makes the use of supervised approaches ineffective.

Furthermore, \ac{ITSR} can be conceptualized as a Markov Decision Problem (MDP). Essentially, an MDP is a mathematical framework that models a decision-making process in a stochastic environment. An MDP consists of a set of states $S$, actions $A$, transition probabilities between states $P(s \mid s', a)$, and a reward function $R(s, a, s')$, with $s,s'\in S$ and $a \in A$, that provides a numerical signal to each state and action. 

Considering ITSR as an MDP, the transformations that can be applied to an image become actions. As such, a trajectory $\mathbf{t}$ is formed by a sequence of those actions. The states of the MDP can be represented as a tuple $(x,x',\mathbf{t})$: the pair of source and target images and the trajectory considered up to that point. For our purpose, the state can be simplified as the pair $(f(x,\mathbf{t}),x')$. Finally, the reward function only returns a positive outcome when the actions of the trajectory transform the source image to the target image; i.e., $f(x,\mathbf{t}) = x'$.

Once the problem is properly formulated as an MDP, RL becomes a suitable paradigm to consider. Note that we can accurately represent the dynamics of the MDP. One can apply a specific transformation to an image at any time and get the result. Also, it is straightforward to know the reward, which is directly related to whether the trajectory gets the target result. Therefore, we will resort to model-based RL. In particular, we consider an approach based on AlphaZero \cite{silver19alphazero}, which learns to estimate sequences of transformations for the input image pairs. In our approach, we train a neural network that learns to retrieve a solution for each input pair through \ac{MCTS}.

\subsection{Monte-Carlo Tree Search}
\ac{MCTS} is a popular model-based RL algorithm applied to stochastic decision-making scenarios. It is a heuristic search that builds a tree that represents the possible outcomes of different actions from a current state. This tree is built incrementally by performing random simulations---often referred to as \emph{rollouts}---and estimating the value of each action in a state from their retrieved outcomes. These estimations are then applied to select the most promising transitions to retrieve a high reward, leading the algorithm to find statistically-based optimal trajectories. \ac{MCTS} is particularly useful when the action space is large and the outcome of each action is uncertain.

AlphaZero \cite{silver19alphazero} combined this algorithm with neural networks to reach more complex scenarios. Specifically, it resorts to an Actor-Critic approach, where the neural network predicts both the probability distribution of the action space---referred to as \emph{actor}---and their value estimation---referred to as \emph{critic}---for a given an input state. This network is trained by building a \ac{MCTS} tree. However, instead of relying on random simulations, the tree is incrementally constructed from network estimations, which are then trained with the retrieved rewards. The main idea is that both actor and critic modules collaborate to find optimal paths in the tree, by correlating values with outcomes and isolating promising actions for each state. In order to balance exploration-exploitation, one can use the Polynomial Upper Confidence Tree (PUCT) score \cite{Auger2013}, which determines the most promising actions in a given state by combining the network estimations with the times they have been taken during training.

\subsection{Transformation Sequence Retrieval}
\label{subsec:retrieval}

After training the neural network using MCTS, we can use the acquired knowledge to retrieve probable sequences for each pair of source and target images. To do this, we build a tree of states in which the most promising action is expanded. The probability of new actions are computed for each node (state) of the tree. These are determined by the probability of a state trajectory---the cumulative probability of its actions---multiplied by the probability that the actor gives to the action in that state. We empirically found that this was more beneficial than considering the value given by the critic to lead the process. The generation of this expansion tree is stopped when a solution is found or a maximum number of nodes is reached.

\section{Experimental setup}
In this section, we define the experimental environment designed to evaluate the proposed methodology. Both the data used and the approaches considered are made available so as to establish a reference benchmark for future attempts.\footnote{\url{https://github.com/emascandela/itsr-mcts}}

\subsection{Network architecture}
\label{sec:framework}
All of the implemented networks trained with \ac{MCTS} follow a generic structure designed to approach the \ac{ITSR} problem. The networks receive the state (source and target images) and provide two outputs, that correspond to the actor and the critic. We divide the process into two modules: a backbone and a decoder.

\begin{figure*}
    \centering
    \includegraphics[width=0.9\textwidth]{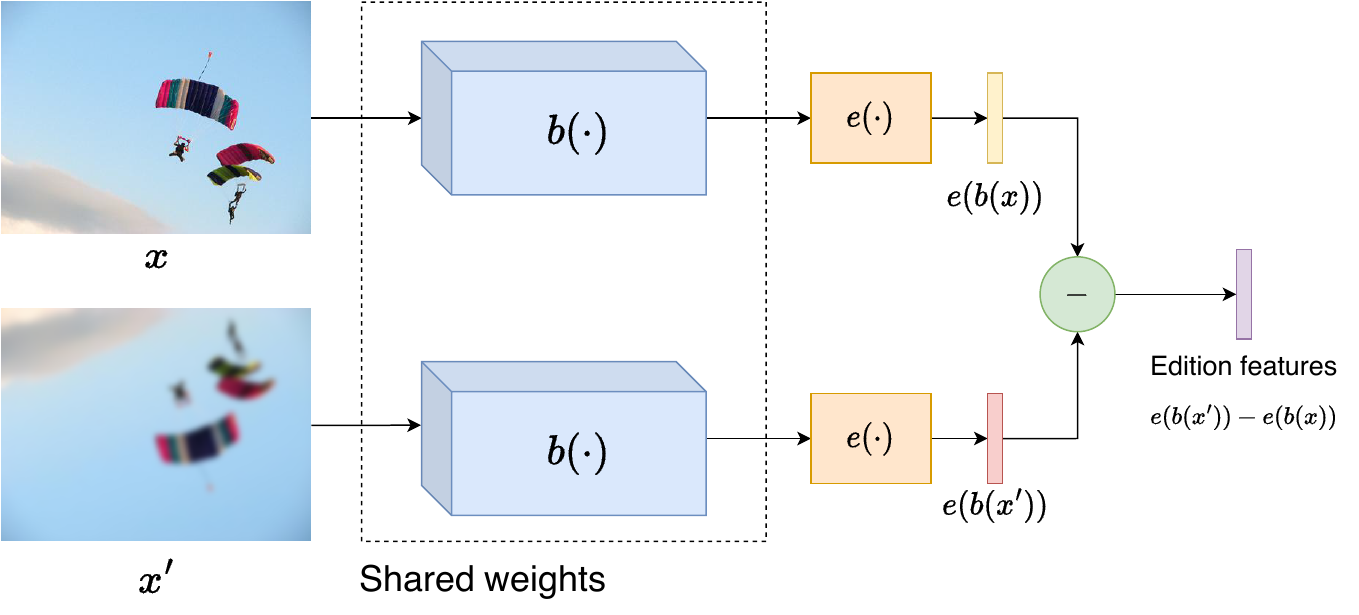}
    \caption{Visualization of the proposed backbone for the edit-features extraction from an input image pair.}
    \label{fig:framework}
\end{figure*}

The first module (backbone) is in charge of generating an appropriate embedding representing the differences between source and target images. This is implemented with a siamese feature extractor, that process both images with the same (shared) weights. Once computed the correspondent feature maps, these are flattened and merged by a subtraction operation. We found empirically that this was the best differentiable operation for this purpose. This also agrees with previous work for comparing images in computer vision \cite{diffnet2020}. A graphic visualization of this module is shown in Fig. \ref{fig:framework}.

The second module (decoder) is in charge of providing the Actor-Critic values from the features of the previous module; that is, both a probability distribution over actions and a value estimation. This is implemented with a single model that first connects the backbone with a dense layer of 128 units, for dimensionality reduction, and then stacks two heads, consisting of one dense layer each, for the actor (softmax activation) and the critic (hyperbolic tangent activation), respectively.

\subsection{Evaluation scenarios}
\label{sec:scenarios}
We evaluate our approach in two different scenarios. First, we measure the performance of a toy problem defined specifically to showcase \ac{ITSR}. Second, we perform an evaluation over a real dataset with common image-processing transformations.

\subsubsection{Toy problem}
To evaluate the suitability of the proposed approach, we define a toy problem that simplifies the image edition feature extraction. This scenario involves the insertion and removal of a set of simple shapes $S$ in a square grid, which has a size of $C \times C$ (see Fig. \ref{fig:problem-example-toy}). We randomly create a pair of images $(x, x')$ and the algorithm determines their transformation sequence. In this work, we consider a grid size $C = 3$ and a set of shapes $S = \{\textrm{Triangle}, \textrm{Circle}, \textrm{Square}\}$. In addition, we consider two setups for the toy problem:

\begin{itemize}
    \item \textbf{Constrained}: Shapes are organized sequentially. Insertions are applied to the first empty slot and removals always refer to the last occupied one. The action space, therefore, consists of the number of figures that can be inserted and a generic removal operation $|A| = |S|+1$. The sequence generation for training is designed to have all removals at the beginning and so there is no multiplicity of possible solutions. This will allow us to draw more interesting insights when compared with other more complex cases.
    \item \textbf{Free}: In this setup, shapes can be inserted or removed freely. Actions also control the slot to which the operation is applied. This significantly increases the action space, since we have an action for each figure and grid position, resulting in $|A| = |S| \times C^2 + C^2$. Even this toy problem has an unmanageable number of solutions.
\end{itemize}

Given that the toy scenario presents low feature recognition requirements, we just implement a LeNet-5 \cite{lenet} model as the feature extraction backbone. This way we focus our analysis on the capabilities of the proposed approach itself.

\subsubsection{Real images}
To bring results in a real-case scenario, we resort to the Imagenette dataset \cite{imagenette}. This corpus consists of a subset of the ImageNet dataset \cite{Deng2009}, which contains $13,394$ samples of the ten most easily classified categories.

In this scenario, we consider a set of five different transformations to be applied, which are rather common in image processing pipelines: $90^o$ rotation, erosion, dilation, blur, and image inversion. To constrain the generation, the sequences have a maximum length of $10$ and the transformations can be repeated only twice. This gives a total number of $326,010$ possible combinations per pair. 

As the real-case scenario presents more complex images, we implemented more complex neural networks as feature extractors for the backbone. Specifically, we implemented three convolutional networks: ResNet18 \cite{Kaiming2016}, EfficientNet-B0 \cite{Tan2019}---referred to as EfN-B0 in the rest of the paper---and ConvNeXt-T \cite{Liu2022}. Recently, Transformer-based architectures are standing as a competing approach to CNN in many computer vision-related tasks. To draw more general conclusions regardless of the specific backbone architecture, we also implemented two Transformer-based models: the Vision Transformer Base \cite{Dosovitskiy2021}---referred to as ViT-B---and the Swin Transformer Network \cite{Liu2021}, which is referred here to as Swin-B.

\subsection{Training}
\label{sec:decoding_imp}
The following details have been considered during training with \ac{MCTS}:

\begin{itemize}
    \item \textbf{Curriculum learning}: To allow learning, we require paths that lead to correct solutions during training time. To do this, \ac{MCTS} has to generate trees where, at least, a leaf node that solves the problem is reached. In the early stages of training, when the network is rather ignorant, this can be difficult to attain. However, this is not inconvenient for short sequences, since just a few iterations are required to find a solution. We, therefore, resort to curriculum learning \cite{bengio09curriculumlearning}, where the neural network first learns to find the solution for short sequences and then progressively trains with longer ones.
    We split the training process into progressive steps, allowing different maximum lengths of transformation sequences in each one.
    
    \item \textbf{Dirichlet Noise}: Neural networks probabilities tend to produce results with tendencies close to $0$ or $1$. Using these probabilities as such may hinder the \ac{MCTS} tree exploration, as it would be highly likely to visit only the nodes promoted implicitly by the critic. To solve this, we introduce noise according to a Dirichlet distribution $\textrm{Dir}(\alpha)$ to the prior probabilities, the same way AlphaZero does \cite{silver19alphazero}. This result distributes the probabilities in favor of the exploration while preserving the distribution estimated by the neural network. The Dirichlet noise is applied to the prior probabilities $p_i$ computed by the neural network with the formula $p_i = p_i \cdot (1 - \epsilon) + \epsilon \cdot \textrm{Dir}(\alpha)$. Here we fix $\alpha$ to $5.0$ and $\epsilon$ to $0.25$.

    \item \textbf{Experience Replay}: In order to improve the training efficiency and stability we introduce the Experience Replay \cite{lin92expreplay} technique. Experience replay stores the agent's experiences (states, actions, and rewards) in a buffer. These experiences are randomly sampled and used during the training process.
\end{itemize}

During the training process, we conduct a total of three curriculum iterations, with each iteration comprising 5, 30, and 50 epochs, respectively. In each of those iterations, we set up different limits on the maximum number of transformations used to generate the ground truth transformation sequence, which are $4$, $8$, and $\infty$ (the last one means that no limit is applied to the ground truth generation). In each epoch, we extract a sample of 1000 trajectories from MCTS, which is executed for 100 iterations for each trajectory. These trajectories are stored in a replay buffer containing 10000 elements. The samples in the replay buffer are used to optimize the neural network's weights via (i) the mean squared error (MSE) loss for the value branch and (ii) the categorical cross-entropy loss (CE) for the policy branch. Within the MCTS algorithm, we utilize a temperature parameter of $\tau=1.0$.

\subsection{Competing baseline}
\label{subsubsec:baseline}
In addition to the proposed \ac{MCTS}, we implemented a supervised approach for comparison purposes. We basically consider the same neural network, yet replace the Actor-Critic heads with a classification layer. 

This network is trained by creating samples of source-target pairs but training the model to classify the next action of the correct sequence. This is done for each step in the sequence, ensuring that all the states of the trajectory are considered during training.

Given that the retrieval process explained in Section \ref{subsec:retrieval} uses the critic (probability distribution over actions), we can keep it using the neural network trained with supervised learning.

\subsection{Additional details}
The following setting has been applied to all the implemented models:
\begin{itemize}
    \item All networks' weights are initialized by following Xavier's uniform distribution \cite{glorot2010understanding}.
    \item Stochastic Gradient Descent is applied as an optimizer, with a momentum of 0.8 and a learning rate of 0.001.
    \item All images are normalized into a range of $[-1, 1]$.
\end{itemize}

\subsection{Evaluation}
Given that it is easy to verify whether a solution is valid, \ac{ITSR} could be solved by simply trying all possible combinations of transformations until finding one that holds the desired condition. However, in interesting cases, there are so many possible combinations that this option becomes infeasible. The goodness of the different approaches, therefore, lies in reaching a valid solution efficiently.

To evaluate our proposal, we consider two metrics. The first is the \emph{single-shot} accuracy, where we only allow the approach to generate one depth-first trajectory to reach a leaf node (terminal state: solution or maximum-allowed depth). In contrast, the tree can generate multiple trajectories, driven by the estimation of the actions learned by the neural network. Considering that each expansion (transformation) has a unit cost, we establish the \emph{Top-K} accuracy, where we measure the ratio of cases that the approach reaches a solution using at most K expansions.

\section{Results}
In this section, we present the results of the different approaches for \ac{ITSR}. For the sake of clarity, we will first analyze the toy problem and then move on to the case with real images.

\subsection{Toy task}
\begin{figure*}[ht]
    \centering
    \includegraphics[width=0.8\textwidth]{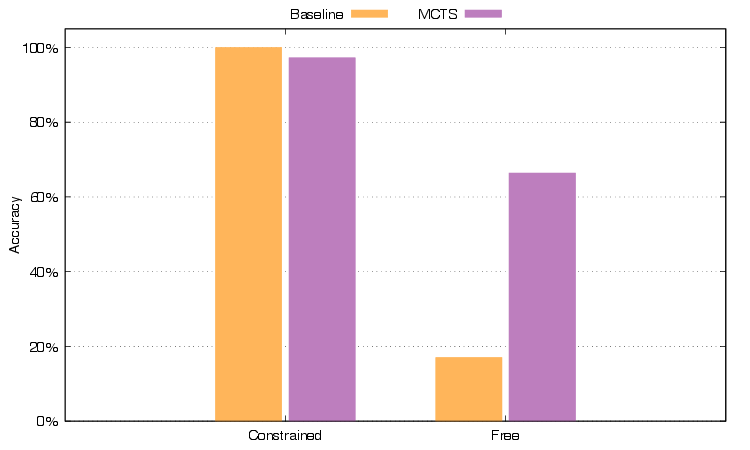}
    
\caption{Accuracy of the single-shot prediction in the toy scenarios: constrained and free.}
    \label{fig:acc-toy}
\end{figure*}

Figure \ref{fig:acc-toy} reports the \emph{single-shot} accuracy of the different alternatives in the toy task.

Concerning the constrained setup, we observe that both approaches succeed in performing \ac{ITSR} (above 95 \%). In this case, the conditions for supervised learning hold, as there is no multiplicity of solutions. Then, the simple baseline approach proves to be not only a successful solution but an improvement over the \ac{MCTS}-based one.

In the free setup, we observe a completely different behavior. While supervised learning is not capable of reporting competitive performance (rate below 20 \%), the \ac{MCTS} approach is capable of learning the task with some success (around 65 \%). This scenario showcases the difficulty of \ac{ITSR} and its associated challenges. Even in a scenario of very low graphical complexity, the multiplicity of solutions and the correlation of the actions of the trajectory make supervised learning ineffective, while RL postulates itself as a good alternative.

\begin{figure*}[ht]
    \centering
    \includegraphics[width=0.8\textwidth]{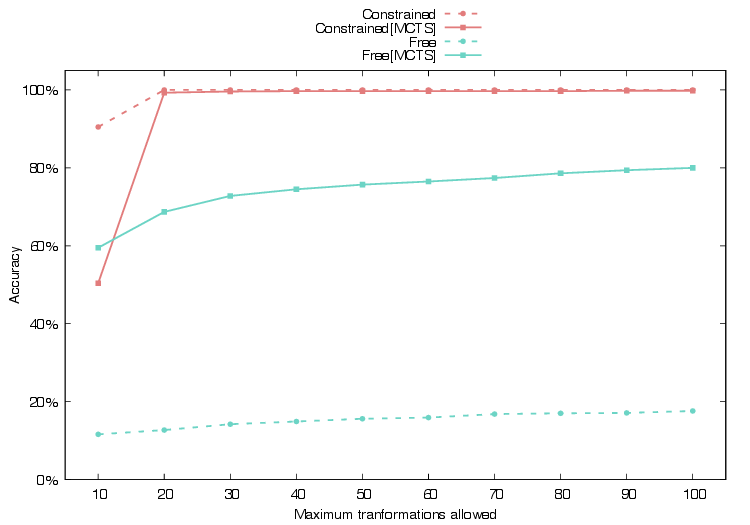}
    
\caption{Top-K accuracy of the evaluation in the toy scenario for different values of $K$.}
    \label{fig:acck-toy}
\end{figure*}

In Fig. \ref{fig:acck-toy} we show the curves that report the \emph{Top-K} accuracy, for values of $K$ in the range between $10$ and $100$. It it observed that the comparison between supervised learning and \ac{MCTS} is basically similar to what was mentioned above for the \emph{single-shot} evaluation. In addition, in this scenario, allowing the models for a greater number of expansions does not bring great benefits. However, it is interesting to note here that the model trained with MCTS has a much lower starting point: this is because its probabilities are sparer and, therefore, by not forcing the depth-first expansion in this evaluation, it often does not reach any leaf node when $K = 10$.

\subsection{Real images}
\begin{figure*}[ht]
    \centering
    \includegraphics[width=0.8\textwidth]{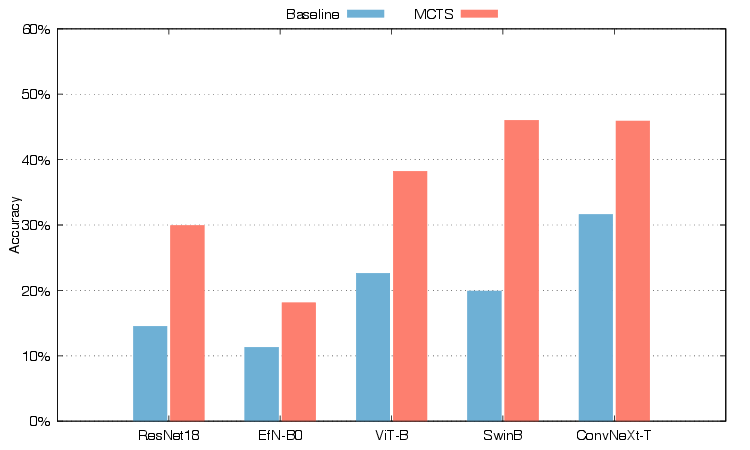}
    
\caption{Accuracy of the single-shot prediction in the real world scenario.}
\label{fig:acc-img}
\end{figure*}

Figure \ref{fig:acc-img} shows the \emph{single-shot} results in the scenario with real images. 

In this case, the complexity of \ac{ITSR} goes up noticeably, so all performances are kept relatively low (below 50 \%). The important remark about these results is to highlight that all the models based on \ac{MCTS} improve the same backbones when trained with supervised learning, except for ResNet-18 which maintains similar performance. While it is true that there is a tendency to improve the results as the capacity of the model increases, this factor is in the background with respect to the learning protocol. This is clearly seen in the deeper models Swin-B and ConvNeXt, which improve from 20 \% to 45 \% and from 32 \% to 44 \%, respectively, by using \ac{MCTS} instead of supervised learning. It is also worth noting that ViT-B with \ac{MCTS} significantly improves these deeper models with supervised learning.

\begin{figure*}[ht]
    \centering
    \includegraphics[width=0.8\textwidth]{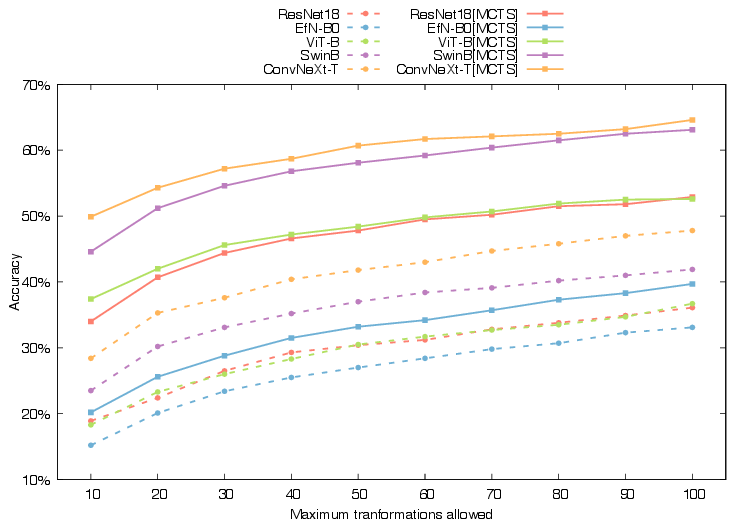}
    \caption{Top-K accuracy of the evaluation in the real scenario for different values of $K$.}
    \label{fig:acck-img}
\end{figure*}

Concerning the \emph{Top-K} accuracy evaluation (see 
Fig.~\ref{fig:acck-img}), we can observe rather general trends for all approaches and implementations. On this occasion, increasing the number of possible transformations entails an increase in accuracy. Although the curves do not reflect a completely linear trend, with $K = 100$ no plateau is still observed. This is an indicator of the good estimation that the models provide, for which each expansion brings them closer to the objective. In this case, this occurs regardless of the learning protocol, although the advantage of the previously discussed \ac{MCTS} approaches remains throughout the range of possibilities.

\begin{figure*}
    \centering
    \begin{subfigure}{0.33\textwidth}
        \includegraphics[width=\linewidth]{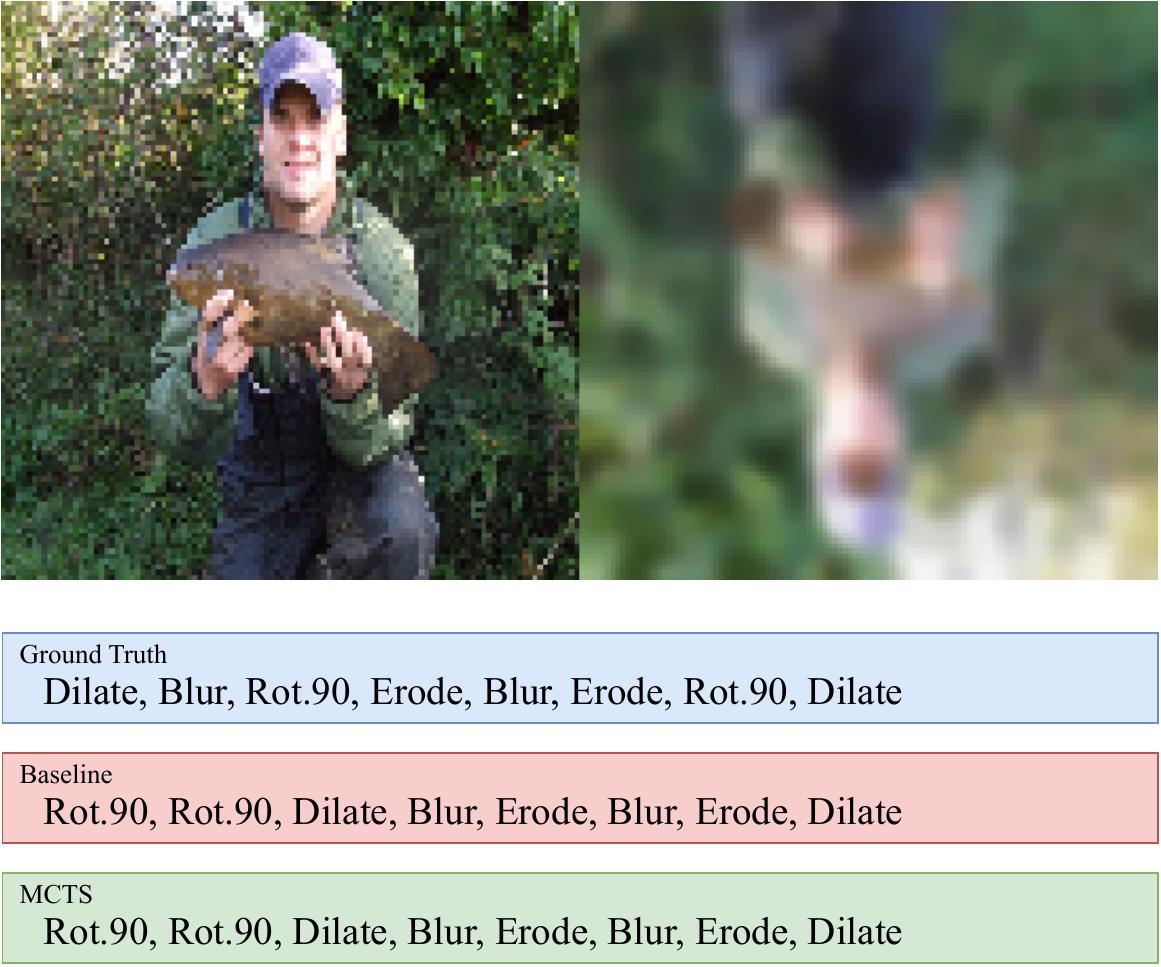}
        \label{fig:sample0}
        \caption{}
    \end{subfigure}
    \begin{subfigure}{0.33\textwidth}
        \includegraphics[width=\linewidth]{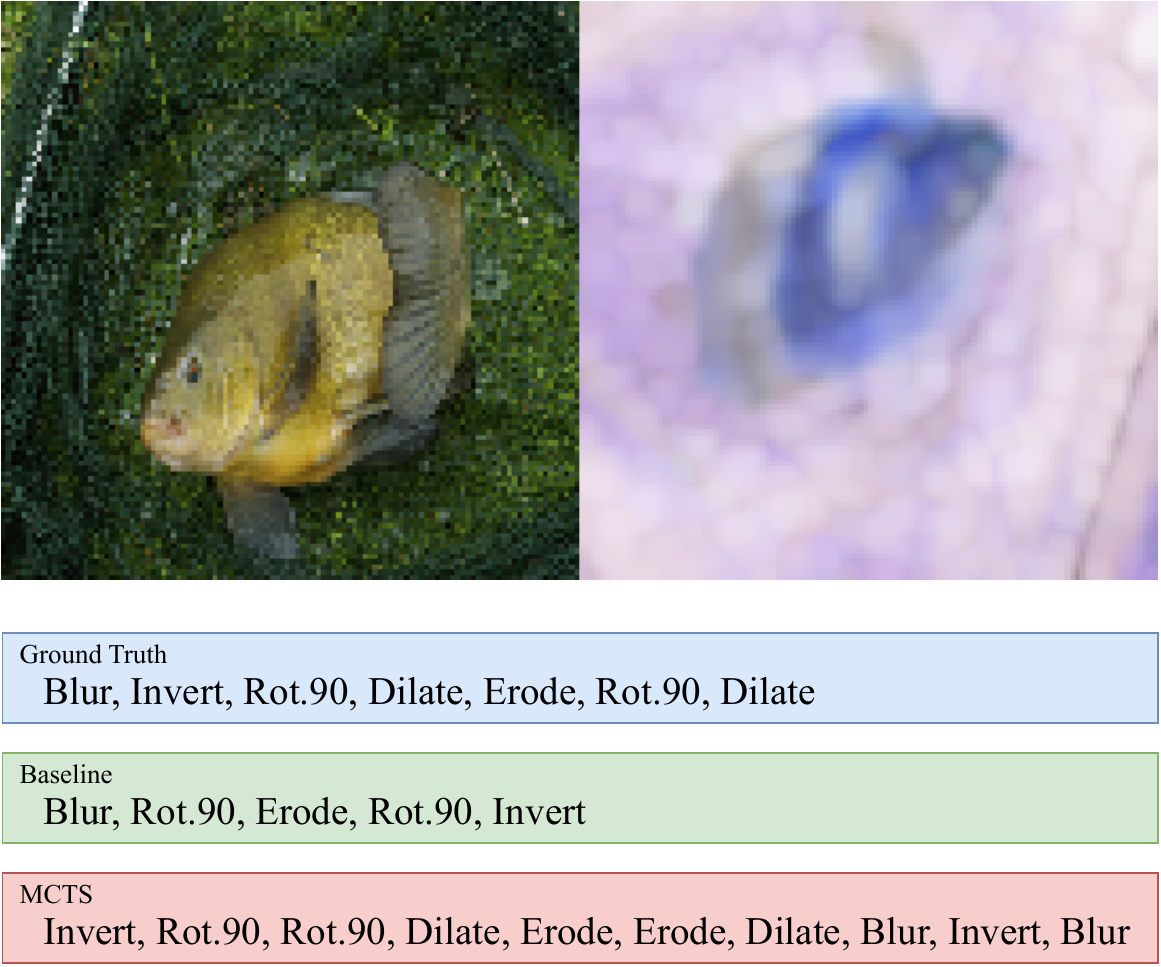}
        \label{fig:sample1}
        \caption{}
    \end{subfigure}
    \begin{subfigure}{0.33\textwidth}
        \includegraphics[width=\linewidth]{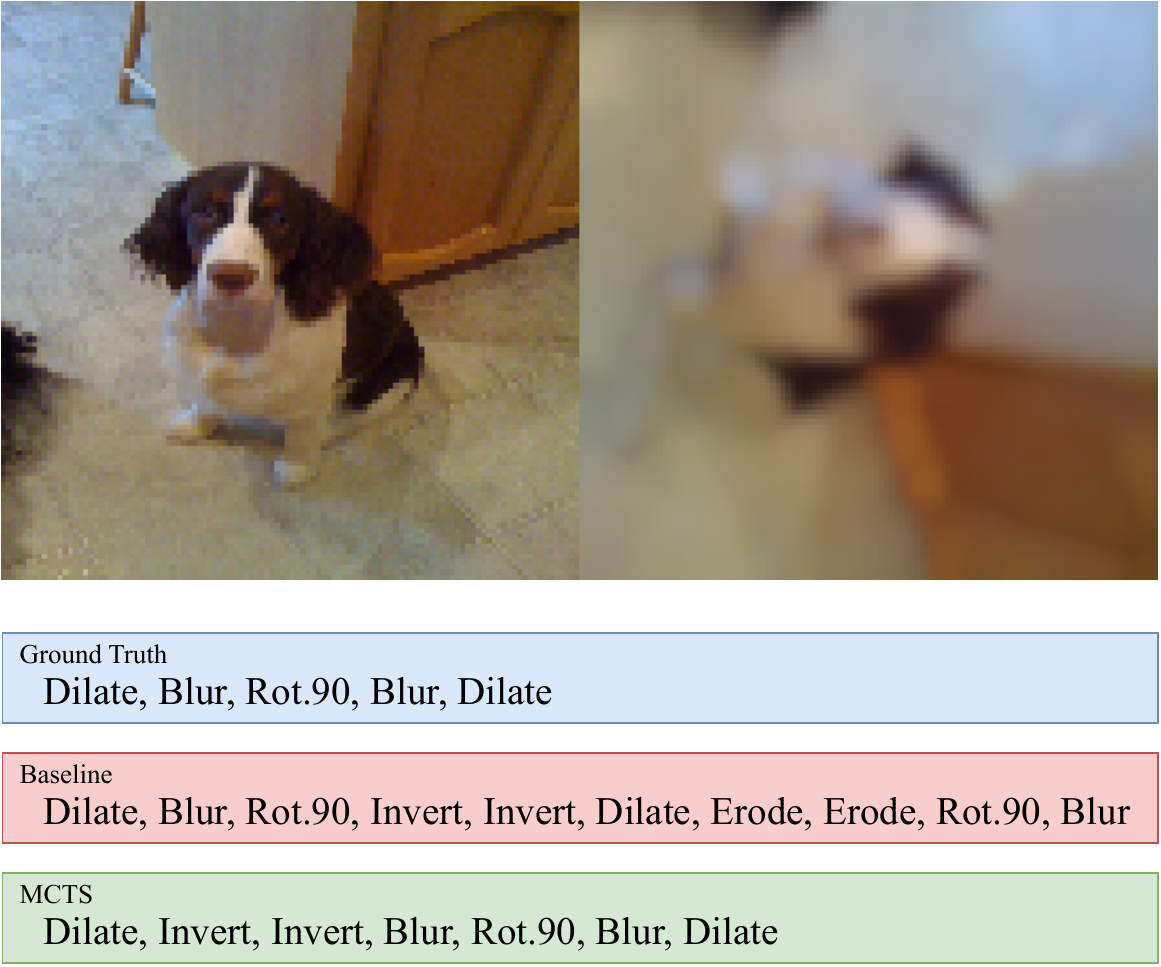}
        \label{fig:sample2}
        \caption{}
    \end{subfigure}
    \begin{subfigure}{0.33\textwidth}
        \includegraphics[width=\linewidth]{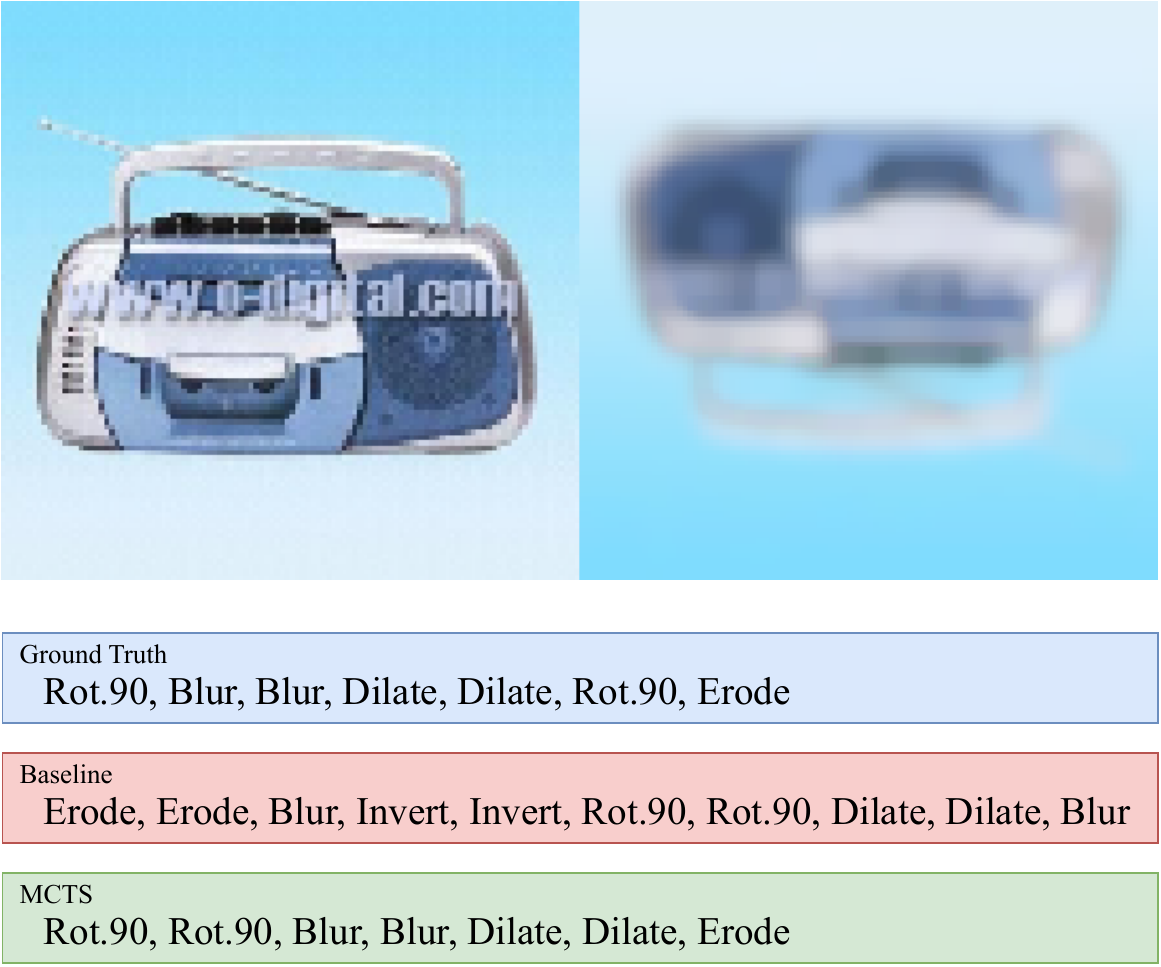}
        \label{fig:sample4}
        \caption{}
    \end{subfigure}
    \begin{subfigure}{0.33\textwidth}
        \includegraphics[width=\linewidth]{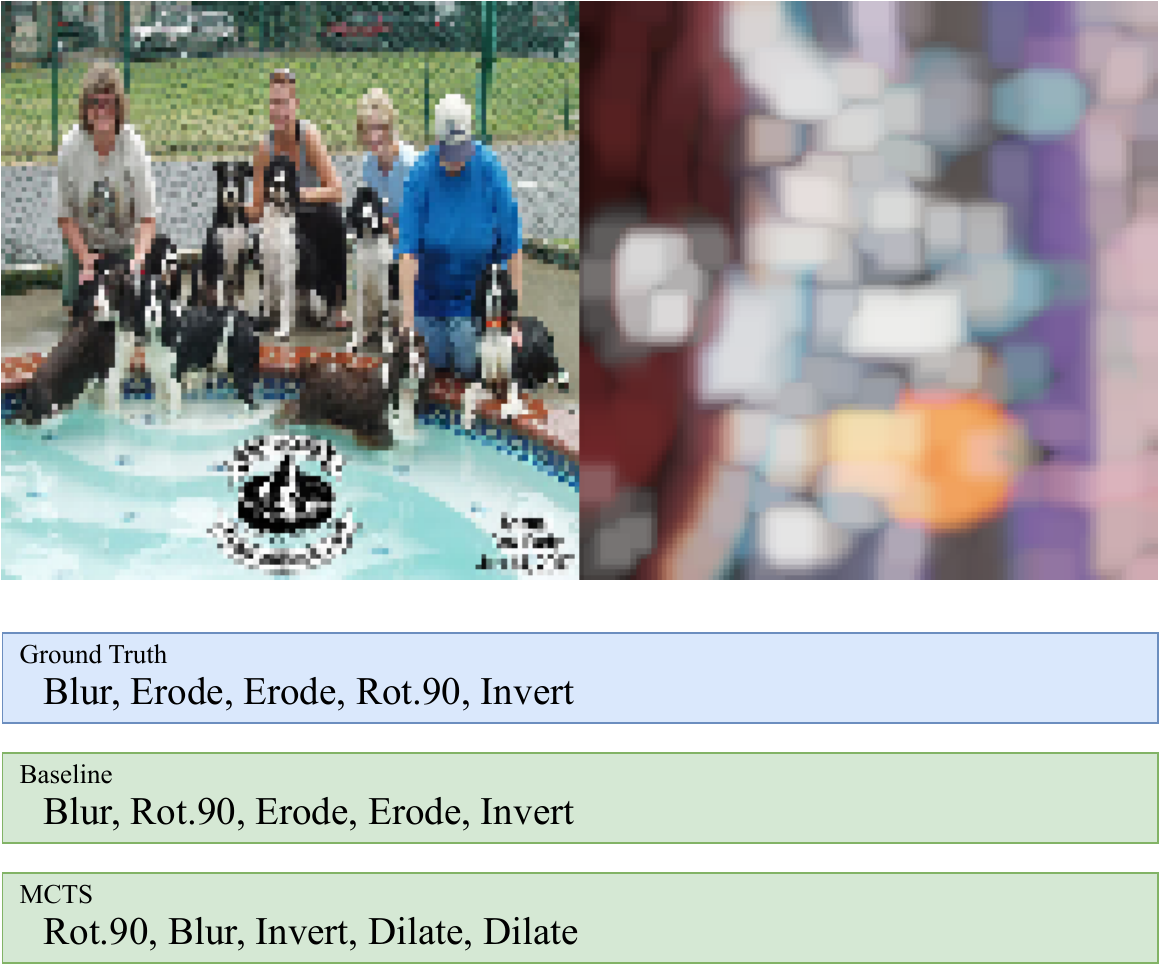}
        \label{fig:sample3}
        \caption{}
    \end{subfigure}
    \begin{subfigure}[b]{0.33\textwidth}
        \includegraphics[width=\linewidth]{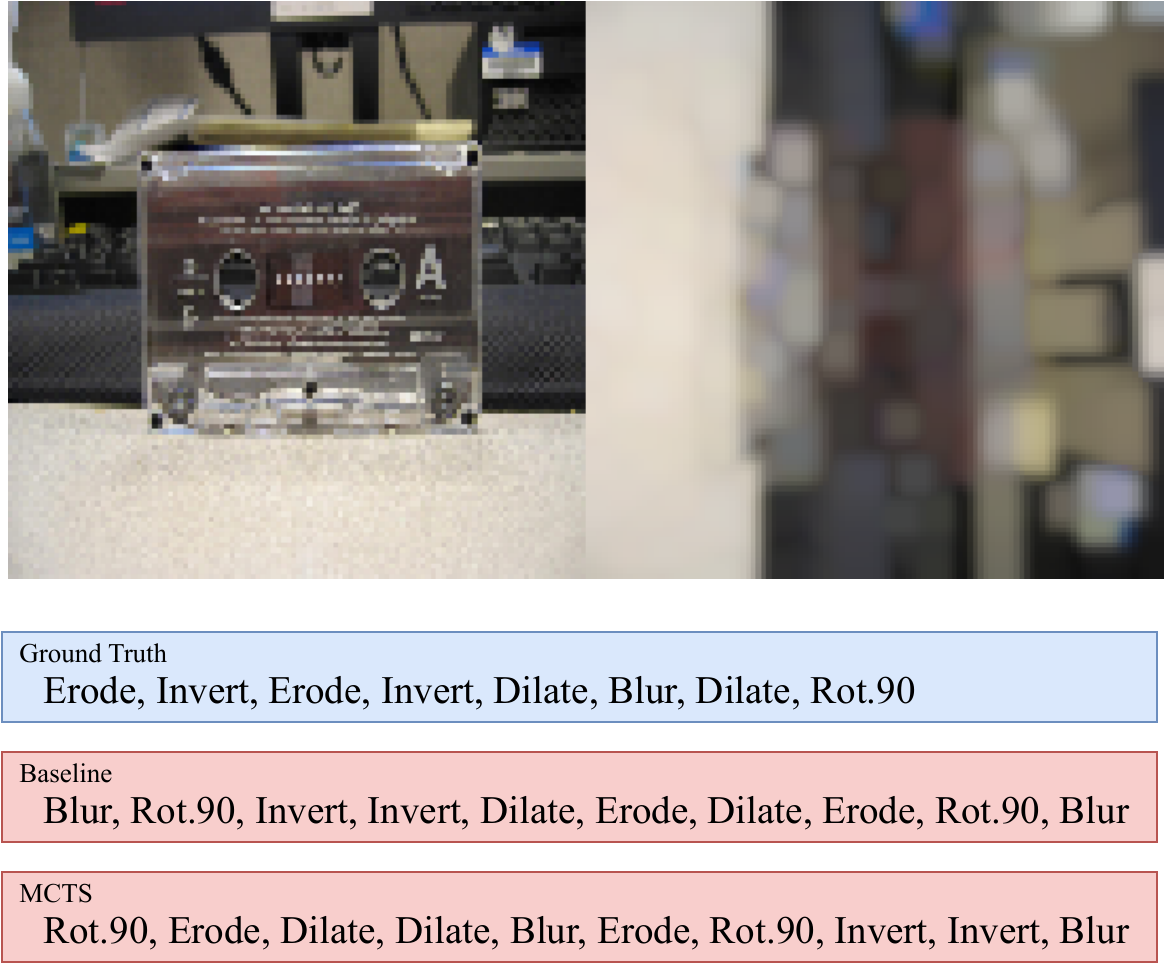}
        \label{fig:sample5}
        \caption{}
    \end{subfigure}
    
    \caption{Examples of the \emph{single-shot} predictions of the baseline and our approach in different samples of the real scenario. The backbone considered is the ConvNeXt. The color in boxes denotes whether the retrieved sequence solves the problem (green) or not (red).}
    \label{fig:output-example}
\end{figure*}

\subsubsection{Qualitative analysis}
To finish with the evaluation of our approach, we have manually inspected the performance of the ConvNeXt model trained with \ac{MCTS} to draw some conclusions:

\begin{itemize}
\item The approach tends to avoid transformations that do not contribute to the output result. For instance, two consecutive inversions, where the second one negates the first one, are not observed.
\item The model often predicts first the transformations whose order has no impact on the final result, such as rotations or inversions. 
\item The approach has difficulties in handling transformations that affect the behavior of other ones. Specifically, sequences that involve inversions and erosions or dilations, such as the sequences $\{\textrm{Invert}, \textrm{Erode}\}$ and $\{\textrm{Dilate}, \textrm{Invert}\}$, which produce the same result.
\item Almost all the samples correctly predicted by the baseline were also correctly predicted by our approach, so we can assume that the supervised approach does not behave better in any particular circumstance than the \ac{MCTS}-based approach.
\end{itemize}

Additionally, in Fig. \ref{fig:output-example} we provide diverse examples of the observed behavior. In addition to a visual clue to understanding the difficulty that \ac{ITSR} entails, with cases that of course humans would find especially complex, it is interesting to note how often the sequence used to generate the sample as ground truth is not the one indicated by the approach, although the result is valid (like in Fig. \ref{fig:output-example}a and \ref{fig:output-example}d).

\section{Conclusions}
In this work, the \ac{ITSR} problem has been addressed for the first time through a model-based RL approach that combines \ac{MCTS} and deep learning. The results showed that: i) \ac{ITSR} is indeed a challenge for existing approaches to computer vision, with wide room for improvement; ii) the RL-based approximation significantly outperforms the supervised setting.

Interesting ways of future work can be considered, such as expanding the scope of the editing operations to other signal domains, dealing with more complex edition languages that go beyond sequences, or combining supervised and reinforcement learning, as has been successfully applied in recent works \cite{fawzi2022discovering}.

\backmatter


\bibliographystyle{sn-mathphys}


\end{document}